\documentclass[10pt,twocolumn,letterpaper]{article}

\usepackage{cvpr}
\usepackage{times}
\usepackage{epsfig}
\usepackage{graphicx}
\usepackage{amsmath}
\usepackage{amssymb}

\usepackage{algorithm}
\usepackage{algorithmic}

\usepackage{setspace}

\usepackage{subfigure}
\usepackage{multirow}
\usepackage{hhline}
\usepackage{enumitem}
\usepackage[T1]{fontenc}

\usepackage[breaklinks=true,bookmarks=false]{hyperref}

\cvprfinalcopy 

\def\cvprPaperID{5591} 

\ifcvprfinal\fi
\begin{document}

\title{Gaussian Temporal Awareness Networks for Action Localization\thanks{{\small This work was performed at JD AI Research.}}}


\author{Fuchen Long$^{\dag}$, Ting Yao$^{\ddag}$, Zhaofan Qiu$^{\dag}$, Xinmei Tian$^{\dag}$, Jiebo Luo$^{\S}$ and Tao Mei$^{\ddag}$ \\
$^{\dag}$University of Science and Technology of China, Hefei, China \\
$^{\ddag}$JD AI Research, Beijing, China \\
$^{\S}$University of Rochester, Rochester, NY USA \\
{\tt\small\{longfc.ustc, tingyao.ustc, zhaofanqiu\}@gmail.com; xinmei@ustc.edu.cn;} \\
{\tt\small\ jluo@cs.rochester.edu; tmei@live.com} \\
}

\maketitle
\thispagestyle{empty}

\begin{abstract}
  Temporally localizing actions in a video is a fundamental challenge in video understanding. Most existing approaches have often drawn inspiration from image object detection and extended the advances, e.g., SSD and Faster R-CNN, to produce temporal locations of an action in a 1D sequence. Nevertheless, the results can suffer from robustness problem due to the design of predetermined temporal scales, which overlooks the temporal structure of an action and limits the utility on detecting actions with complex variations. In this paper, we propose to address the problem by introducing Gaussian kernels to dynamically optimize temporal scale of each action proposal. Specifically, we present Gaussian Temporal Awareness Networks (GTAN) --- a new architecture that novelly integrates the exploitation of temporal structure into an one-stage action localization framework. Technically, GTAN models the temporal structure through learning a set of Gaussian kernels, each for a cell in the feature maps. Each Gaussian kernel corresponds to a particular interval of an action proposal and a mixture of Gaussian kernels could further characterize action proposals with various length. Moreover, the values in each Gaussian curve reflect the contextual contributions to the localization of an action proposal. Extensive experiments are conducted on both THUMOS14 and ActivityNet v1.3 datasets, and superior results are reported when comparing to state-of-the-art approaches. More remarkably, GTAN achieves 1.9\% and 1.1\% improvements in mAP on testing set of the two datasets.
\end{abstract}

\section{Introduction}
With the tremendous increase of online and personal media archives, people are generating, storing and consuming a large collection of videos. The trend encourages the development of effective and efficient algorithms to intelligently parse video data. One fundamental challenge that underlies the success of these advances is action detection in videos from both temporal aspect \cite{Gaidon:PAMI13,Geest:ECCV16,Lea:CVPR17,Shou:CVPR16,yao2017msr,Xiong:ICCV17} and spatio-temporal aspect \cite{Gkioxari:CVPR15,Dong:ECCV18}. In this work, the main focus is temporal action detection/localization, which is to locate the exact time stamps of the starting and the ending of an action, and recognize the action with a set of categories.

\begin{figure}[!tb]
         \centering\includegraphics[width=0.42\textwidth]{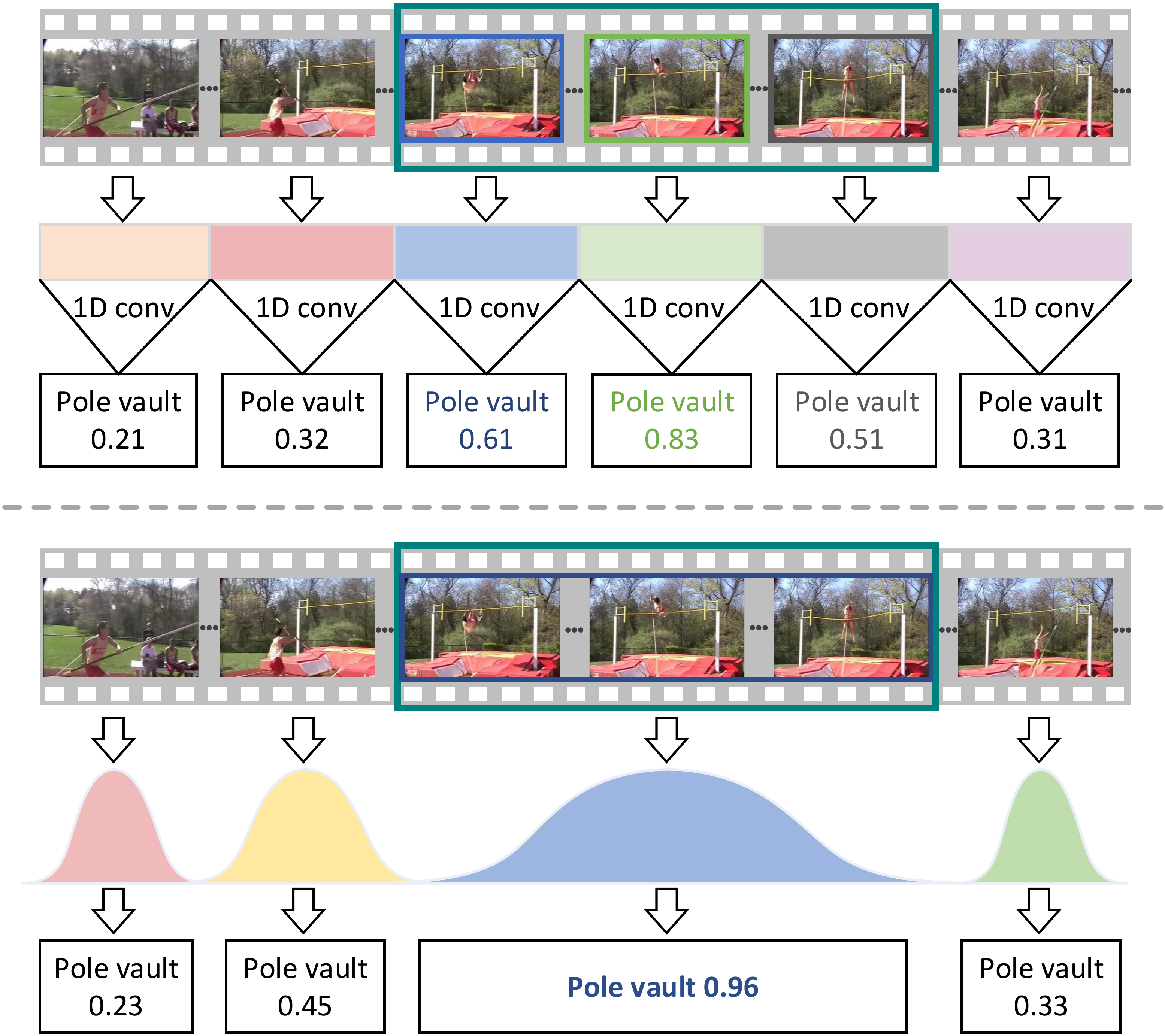}
         \caption{\small The intuition of a typical one-stage action localization (upper) and our GTAN (lower). The typical method fixes temporal scale in each feature map and seldom explores temporal structure of an action. In contrast, temporal structure is taken into account in our GTAN through learning a set of Gaussian kernels.}
         \label{fig1:1}
   \vspace{-0.24in}
\end{figure}

One natural way of temporal action localization is to extend image object detection frameworks, e.g., SSD \cite{Liu:ECCV16} or Faster R-CNN \cite{Ren:NIPS15}, for producing spatial bounding boxes in a 2D image to temporal localization of an action in a 1D sequence \cite{Chao:CVPR18,Lin:MM17}. The upper part of Figure \ref{fig1:1} conceptualizes a typical process of one-stage action localization. In general, the frame-level or clip-level features in the video sequence are first aggregated into one feature map, and then multiple 1D temporal convolutional layers are devised to increase the size of temporal receptive fields and predict action proposals. However, the temporal scale corresponding to the cell in each feature map is fixed, making such method unable to capture the inherent temporal structure of an action. As such, one ground-truth action proposal in the green box is detected as three ones in this case. Instead, we propose to alleviate the problem by exploring the temporal structure of an action through learning a Gaussian kernel for each cell, which dynamically indicates a particular interval of an action proposal. A mixture of Gaussian kernels could even be grouped to describe an action, which is more flexible to localize action proposals with various length as illustrated in the bottom part of Figure \ref{fig1:1}. More importantly, the contextual information is naturally involved with the feature pooling based on the weights in Gaussian curve.

By delving into temporal structure of an action, we present a novel Gaussian Temporal Awareness Networks (GTAN) architecture for one-stage action localization. Given a video, a 3D ConvNet is utilized as the backbone to extract clip-level features, which are sequentially concatenated into a feature map. A couple of convolutional layers plus max-pooling layer are firstly employed to shorten the feature map and increase the temporal size of receptive fields. Then, a cascaded of 1D temporal convolutional layers (anchor layers) continuously shorten the feature map and output anchor feature map, which consists of features of each cell (anchor). On the top of each anchor layer, a Gaussian kernel is learnt for each cell to dynamically predict a particular interval of an action proposal corresponding to that cell. Multiple Gaussian kernels could even be mixed to capture action proposals with arbitrary length. Through Gaussian pooling, the features of each cell is upgraded by aggregating the features of contextual cells weighted by the values in the Gaussian curve for final action proposal prediction. The whole architecture is end-to-end optimized by minimizing one classification loss plus two regression losses, i.e., localization loss and overlap loss.

The main contribution of this work is the design of an one-stage architecture GTAN for addressing the issue of temporal action localization in videos. The solution also leads to the elegant view of how temporal structure of an action should be leveraged for detecting actions with various length and how contextual information should be utilized for boosting temporal localization, which are problems not yet fully understood in the literature.

\begin{figure*}[!tb]
	\centering\includegraphics[width=0.88\textwidth]{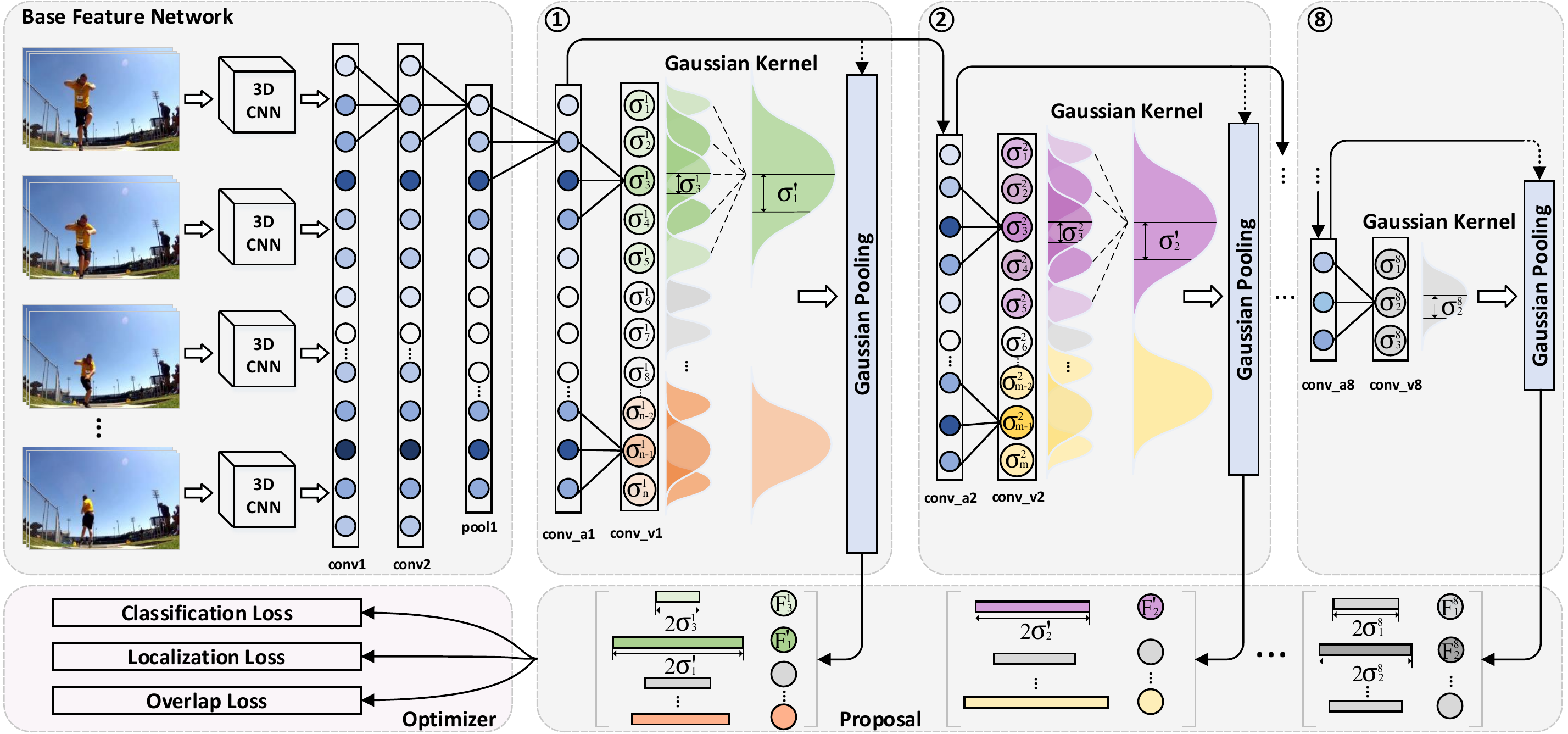}
    \vspace{-0.03in}
	\caption{\small An overview of our Gaussian Temporal Awareness Networks (GTAN) architecture. The input video is encoded into a series of clip-level features via a 3D ConvNet, which are sequentially concatenated as a feature map. Two 1D convolutional layers plus one max-pooling layer are followed to increase the temporal size of receptive fields. Eight 1D convolutional layers are cascaded to generate multiple feature maps in different temporal resolution. On the top of each feature map, a Gaussian kernel is learnt on each cell to predict a particular interval of an action proposal. Moreover, multiple Gaussian kernels with high overlap are mixed to a larger one for detecting long actions with various length. Through Gaussian pooling, the action proposal is generated by aggregating the features of contextual cells weighted by the values in the Gaussian curve. The GTAN is jointly optimized with action classification loss plus two regression losses, i.e., localization loss and overlap loss for each proposal. Better viewed in original color pdf.}
\label{fig2:1}
\vspace{-0.22in}
\end{figure*}

\section{Related Work}
We briefly group the related works into two categories: temporal action proposal and temporal action detection. The former focuses on investigating how to precisely localize video segments which contain actions, while the latter further classifies these actions into known classes.

We summarize the approaches on temporal action proposal mainly into two directions: content-independent proposal and content-dependent proposal. The main stream of content-independent proposal algorithms is uniformly or sliding window-ly sampling in a video \cite{Oneata:ICCV13,Tang:ICCV13,Yuan:CVPR16}, which leads to huge computations for further classification. In contrast, content-dependent proposal methods, e.g., \cite{Buch:CVPR17,Escorcia:ECCV16,Gao:ECCV18,Gao:ICCV17,Lin:ECCV18}, utilize the label of action proposals during training. For instance, Escorcia \emph{et al.}~\cite{Escorcia:ECCV16} leverage Long Short-Term Memory cells to learn an appropriate encoding of a video sequence as a set of discriminative states to indicate proposal scores. Though the method avoids running sliding windows of multiple scales, there is still the need of executing an overlapping sliding window that is inapplicable when the video duration is long. To address this problem, Single Stream Temporal proposal (SST) \cite{Buch:CVPR17} generates proposals with only one single pass by utilizing a recurrent GRU-based model, and Temporal Unit Regression Network (TURN) \cite{Gao:ICCV17} builds video units in a pyramid manner to avoid window overlapping. Different from the above methods which generate proposals in a fixed multi-scale manner, Boundary Sensitive Network (BSN) \cite{Lin:ECCV18} localizes the action boundaries based on three actionness curves in a more flexible way. Nevertheless, such actionness-based methods may fail in locating dense and short actions because of the difficulty to discriminate between very close starting and ending peaks in the curve.

Once the localization of action proposals completes, the natural way for temporal action detection is to further classify the proposals into known action classes, making the process in two-stage manner \cite{Chao:CVPR18,Heilbron:CVPR17,Shou:CVPR17,Shou:CVPR16,Xu:ICCV17,Xiong:ICCV17}. However, the separate of proposal generation and classification may result in sub-optimal solutions. To further facilitate temporal action detection, there have been several one-stage techniques \cite{Buch:BMVC17,Lin:MM17,Yeung:CVPR16} being proposed recently. For example, Single Stream Temporal Action Detection (SS-TAD)~\cite{Buch:BMVC17} utilizes the Recurrent Neural Network (RNN) based architecture to jointly learn action proposal and classification. Inspired by SSD~\cite{Liu:ECCV16}, Lin \emph{et al.}~\cite{Lin:MM17} devise 1D temporal convolution to generate multiple temporal action anchors for action proposal and detection. Moreover, with the development of reinforcement learning, Yeung \emph{et al.}~\cite{Yeung:CVPR16} explore RNN to learn a glimpse policy for predicting the starting and ending points of actions in an end-to-end manner. Nevertheless, most of one-stage methods are still facing the challenge in localizing all the action proposals due to the predetermined temporal scales.

In short, our approach belongs to one-stage temporal action detection techniques. Different from the aforementioned one-stage methods which often predetermine temporal scales of action proposals, our GTAN in this paper contributes by studying not only learning temporal structure through Gaussian kernels, but also how the contextual information can be better leveraged for action localization.

\section{Gaussian Temporal Awareness Networks}
In this section we present the proposed Gaussian Temporal Awareness Networks (GTAN) in detail. Figure~\ref{fig2:1} illustrates an overview of our architecture for action localization. It consists of two main components: a base feature network and a cascaded of 1D temporal convolutional layers with Gaussian kernels.
The base feature network is to extract feature map from sequential video clips, which will be fed into cascaded 1D convolutional layers to generate multiple feature maps in different temporal resolution.
For each cell in one feature map, a Gaussian kernel is learnt to control temporal scale of an action proposal corresponding to that cell as training proceeds. 
Furthermore, a Gaussian Kernel Grouping algorithm is devised to merge multiple Gaussian kernels with high overlap to a larger one for capturing long actions with arbitrary length. Specifically, each action proposal is generated by aggregating the features of contextual cells weighted by the values in the Gaussian curve.
The whole network is jointly optimized with action classification loss plus two regression losses, i.e., localization loss and overlap loss, which are utilized to learn action category label, default temporal boundary adjustment and overlap confidence score for each action proposal, respectively.

\subsection{Base Feature Network}
The ultimate target of action localization is to detect action instances in temporal dimension. Given an input video, we first extract clip-level features from continuous clips via a 3D ConvNet which could capture both appearance and motion information of the video.
Specifically, a sequence of features \mbox{$\{f_i\}_{i=0}^{T-1}$} are extracted from 3D ConvNet, where $T$ is the temporal length. We concatenate all the features into one feature map and then feed the map into two 1D convolutional layers (``conv1'' and ``conv2'' with temporal kernel size 3, stride 1) plus one max-pooling layer (``pool1'' with temporal kernel size 3, stride 2) to increase the temporal size of receptive fields. The base feature network is composed of 3D ConvNet, two 1D convolutional layers and max-pooling layer. The outputs of the base feature network are further exploited for action proposal generation.

\subsection{Gaussian Kernel Learning} \label{sec:3.2}
Given the feature map output from the base feature network, a natural way for one-stage action localization is to stack 1D temporal convolutional layers (anchor layers) to generate proposals (anchors) for classification and boundary regression.
This kind of structure with predetermined temporal scale in each anchor layer can capture action proposals whose temporal intervals are well aligned with the size of receptive fields, however, posts difficulty to the detection of proposals with various length. The design limits the utility on localizing actions with complex variations.

To address this issue, we introduce temporal Gaussian kernel to dynamically control the temporal scales of proposals in each feature map. In the literature, there has been evidences on the use of Gaussian kernels for event detection in videos \cite{Piergiovanni:AAAI17, Piergiovanni:CVPR18}.
In particular, as shown in Figure \ref{fig2:1}, eight 1D temporal convolutional layers (anchor layers) are first cascaded for action proposal generation in different temporal resolution. For each cell in the feature map of the anchor layer, a Gaussian kernel is learnt to predict a particular interval of an action proposal corresponding to that cell.
Formally, we denote the feature map of $j$-th convolutional layer as \mbox{$\{f_i\}_{i=0}^{T^j-1}\in{\mathbb{R}^{T^j\times D^j}}$}, $1\leq j\leq 8$, where $T^j$ and $D^j$ are the temporal length and feature dimension of the feature map. For a proposal $P^j_{t}$ whose center location is $t$, we leverage its temporal scale by a Gaussian kernel $G^j_{t}$.
The standard deviation $\sigma^j_{t}$ of $G^j_{t}$ is learnt via a 1D convolutional layer on a $3 \times D^j$ feature map cell, and the value is constrained within the range $(0,1)$ through a sigmoid operation. The weights of the Gaussian kernel $G^j_{t}$ are defined~as
\begin{equation}\label{Eq2:1}\small
\begin{split}
&W^j_t[i] = \frac{1}{Z}\exp(-\frac{(p_i-\mu_t)^2}{2{\sigma^j_{t}}^2})~, \\
&s.t.~~~p_i = \frac{i}{T^j},~ \mu_t = \frac{t}{T^j}~,\\
&~~~~~~~~i \in \{0,1,...,T^j-1\},~~~~ t \in \{0,1,...,T^j-1\},
\end{split}
\end{equation}
where $Z$ is the normalizing constant.
Taking the spirit from the theory that the $\sigma^j_{t}$ could be considered as a measure of width (Root Mean Square width, RMS) in Gaussian kernel $G^j_{t}$, we utilize $\sigma^j_{t}$ as the interval measure of action proposal $P^j_{t}$.
Specifically, the $\sigma^j_{t}$ can be multiplied with a certain ratio to represent the default temporal boundary:
\begin{equation}\label{Eq2:3}\small
a_c = (t+0.5)/T^{j},~~~~a_w = r_d \cdot 2\sigma^j_{t}/T^{j},
\end{equation}
where $a_c$ and $a_w$ are the center location and width of default temporal boundary and $r_d$ represents temporal scale ratio. The $W^j_t$ is also utilized for feature aggregation with a pooling mechanism to generate action proposals, which will be elaborated in Section \ref{sec:3.4}.

\begin{figure}[!tb]
	\centering\includegraphics[width=0.34\textwidth]{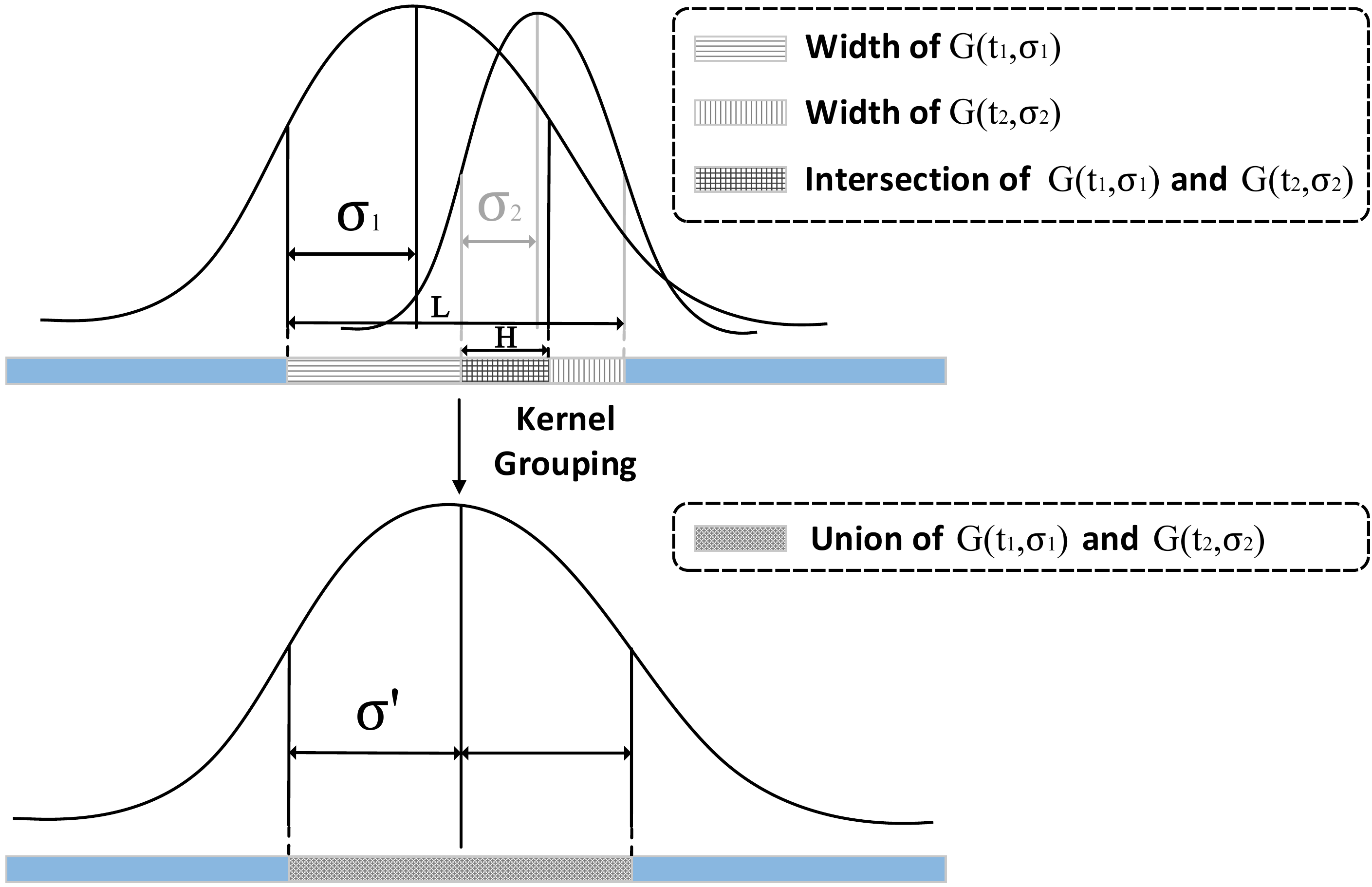}
	\caption{\small Visualization of Gaussian Kernel Grouping.}
	\label{fig2:2}
   \vspace{-0.25in}
\end{figure}

Compared to the conventional 1D convolutional anchor layer which fixes the temporal scale as $1/T^j$ in $j$-th layer, ours employs the dynamic temporal scales by leveraging the learned Gaussian kernel of each proposal to explore the action instances with complex variations.

\subsection{Gaussian Kernel Grouping}

\begin{algorithm}[htb]\small
       \caption{\small Gaussian Kernel Grouping}
       \label{GKG}
       {\fontsize{7.5}{7.5}\selectfont
       \begin{algorithmic}[1]
              \REQUIRE ~~\\
              Original Gaussian kernel set $\mathbb{S} = \{G(t_i, \sigma_i)\}^{T-1}_{i=0}$;\\
              Intersection over Union (IoU) threshold $\varepsilon$; \\
              \ENSURE ~~\\
              Mixed Gaussian kernel set $\mathbb{G}$; \\
              \STATE Choose the beginning grouping position $p = 0$;
              \STATE Initialize mixed Gaussian kernel set $\mathbb{G} = \varnothing$;
              \STATE Initialize base Gaussian kernel $G_{bs} = G(t_p, \sigma_p)$, the ending grouping position $z = p+1$;
              \WHILE{$p \leq T-1$}
              \STATE Compute IoU value $O$ between kernel $G_{bs}$ and $G(t_z, \sigma_z)$;
              \IF{$O > \varepsilon$}
              \STATE Group $G_{bs}$ and $G(t_z, \sigma_z)$ to $G'$ according to Eq.(\ref{Eq2:4}), replace $G_{bs}$ with the new mixed kernel $G'$;
              \ELSE
              \STATE Add kernel $G_{bs}$ to mixed kernel set $\mathbb{G}$;
              \STATE $p = z$,~~~$G_{bs} = G(t_p, \sigma_p)$;
              \ENDIF
              \STATE $z = z + 1$;
              \ENDWHILE
              \RETURN $\mathbb{G}$
       \end{algorithmic}
      }
\end{algorithm}

Through learning temporal Gaussian kernels, the temporal scales of most action instances can be characterized with the predicted standard deviation.
However, if the learned Gaussian kernels span and overlap with each other, that may implicitly indicate a long action centered at a flexible position among these Gaussian kernels. In other words, utilizing the center locations of these original Gaussian kernels to represent this long proposal may not be appropriate.
To alleviate this issue, we attempt to generate a set of new Gaussian kernels to predict center location and temporal scales of proposals for long action.
Inspired by the idea of temporal actionness grouping in~\cite{Xiong:ICCV17}, we propose a novel Gaussian Kernel Grouping algorithm for this target.

Figure~\ref{fig2:2} illustrates the process of temporal Gaussian Kernel Grouping. Given two adjacent Gaussian kernels $G(t_1, \sigma_1)$ and $G(t_2, \sigma_2)$ whose center location and standard deviation are $t$ and $\sigma$, we compute the temporal intersection and union between two kernels by using the width $a_w$ of the default temporal boundary defined in Section \ref{sec:3.2}.
In upper part of Figure~\ref{fig2:2}, the length of temporal intersection between two kernels is $H$, while the length of union is $L$. 
If the Intersection over Union (IoU) between the two kernels $H/L$ exceeds a certain threshold $\varepsilon$, we merge them into one Gaussian kernel (bottom part of Figure~\ref{fig2:2}). The new mixed Gaussian kernel is formulated as follows
\begin{equation}\label{Eq2:4}\small
\begin{split}
&W[i] = \frac{1}{Z}\exp(-\frac{(p_i-\mu')^2}{2{\sigma'}^2})~, \\
&s.t.~~~ p_i = \frac{i}{T}, ~~~ \mu' = \frac{t_1+t_2}{2 \cdot T}~, ~~~ \sigma' = \frac{L}{2}~,\\
&~~~~~~~~i \in \{0,1,...,T-1\}.
\end{split}
\end{equation}

In each feature map, Algorithm~\ref{GKG} details the grouping steps to generate merged kernels.

\subsection{Gaussian Pooling}\label{sec:3.4}
\begin{figure}[!tb]
	\centering\includegraphics[width=0.44\textwidth]{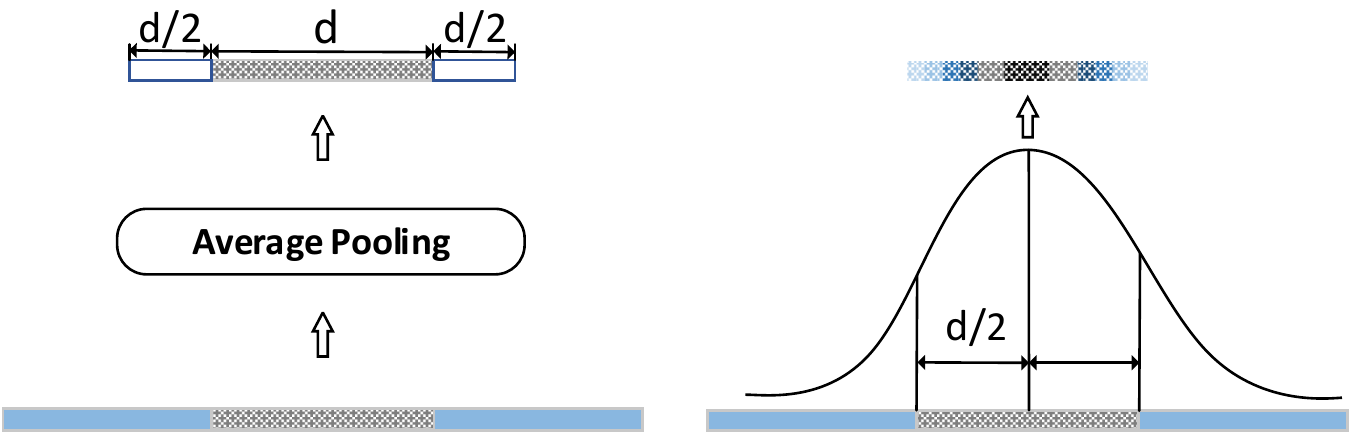}
    \vspace{-0.06in}
	\caption{\small Comparisons of manual extension plus average-pooling strategy (left) and Gaussian pooling strategy (right) for involving temporal contextual information of action proposals.}
	\label{fig2:3}
   \vspace{-0.18in}
\end{figure}

With the learned and mixed Gaussian kernels, we calculate the weighted sum of the feature map based on the values in Gaussian curve and obtain the aggregated feature $F$.
Specifically, given the weighting coefficients $W^j_{t}$ of Gaussian kernel $G^j_{t}$ at center location $t$ in $j$-th layer, the aggregated feature for proposal $P^j_{t}$ is formulated as
\begin{equation}\label{Eq2:2}\small
\begin{split}
&F^j_t = \frac{1}{T^j}\sum\nolimits_{i=0}^{T^j-1} W^j_t[i] \cdot f_i, \\
\end{split}
\end{equation}
where the representation $F^j_{t}$ is further exploited for the action classification and temporal boundary regression.

The above Gaussian pooling mechanism inherently takes the contextual contributions around each action proposal into account.
In contrast to the manual extension plus average-pooling strategy to capture video context information (left part of Figure~\ref{fig2:3}), ours provides an elegant alternative to adaptively learn the weighted representation (right part of Figure~\ref{fig2:3}) based on the importance.

\subsection{Network Optimization}

Given the representation of each proposal from Gaussian pooling, three 1D convolutional layers are utilized in parallel to predict action classification scores, localization parameters and overlap parameter, respectively. Action classification scores $\mathbf{y^a}=[y^a_0, y^a_1, ..., y^a_{C}]$ indicate the probabilities belonging to $C$ action classes plus one ``background'' class. Localization parameters $(\Delta c, \Delta w)$ denote temporal offsets relative to default center location $a_c$ and width $a_w$, which are leveraged to adjust the temporal coordinate
\begin{equation}\label{Eq2:6}\small
       \begin{split}
              \varphi_c = a_c + \alpha_1 a_w \Delta c ~~~{\rm{and}} ~~~\varphi_w = a_w \exp{(\alpha_2 \Delta w)}~,
       \end{split}
\end{equation}
where \mbox{$\varphi_c$}, \mbox{$\varphi_w$} are refined center location and width of the proposal. The $\alpha _1$, $\alpha _2$ are utilized to control the impact of temporal offsets.
In particular, we define an overlap parameter $y_{ov}$ to represent the precise IoU prediction of the proposal, which benefits the proposal re-ranking in prediction.

In the training stage, we accumulate all the proposals from Gaussian pooling and produce the action instances through prediction layer. The overall training objective in our GTAN is formulated as a multi-task loss by integrating action classification loss ($L_{cls}$) and two regression losses, i.e., localization loss ($L_{loc}$) and overlap loss ($L_{ov}$):
\begin{equation}\label{Eq2:9}
       \small
       {L} = {L}_{cls} + \beta {L}_{loc} + \gamma {L}_{ov},
\end{equation}
where \mbox{$\beta$} and \mbox{$\gamma$} are the trade-off parameters. Specifically, we measure the classification loss $L_{cls}$ via the softmax loss:
\begin{equation}\label{Eq2:5}\small
L_{cls} = -\sum\limits_{n=0}^{C}I_{n=c}\log(y^a_{n}),
\end{equation}
where indicator function $I_{n=c}=1$ if $n$ equals to ground truth action label $c$, otherwise $I_{n=c}=0$.
We denote $g_{iou}$ as the IoU between default temporal boundary of this proposal and its corresponding closest ground truth. If the \mbox{$g_{iou}$} of this proposal is larger than $0.8$, we set it as a foreground sample. If \mbox{$g_{iou}$} is lower than $0.3$, it will be set as background sample.
The ratio between foreground and background samples is set as 1.0 during training.
The localization loss is devised as Smooth L1 loss \cite{Girshick:ICCV15} (\mbox{$S_{L1}$}) between the predicted foreground proposal and the closest ground truth instance of the proposal, which is computed by
\begin{equation}\label{Eq2:7}
       \small
       {L}_{loc} = S_{L1}(\varphi_{c}-g_{c})+S_{L1}(\varphi_{w}-g_{w}),
\end{equation}
where \mbox{$g_{c}$} and \mbox{$g_{w}$} represents the center location and width of the proposal's closest ground truth instance, respectively.
For overlap loss, we adopt the mean square error (MSE) loss to optimize it as follows:
\begin{equation}\label{Eq2:8}
\small
L_{ov} = (y_{ov} - g_{iou})^2.
\end{equation}

Eventually, the whole network is trained in an end-to-end manner by penalizing the three losses.

\subsection{Prediction and Post-processing}
During prediction of action localization, the final ranking score $y_f$ of each candidate action proposal depends on both action classification scores $\mathbf{y^a}$ and overlap parameter $y_{ov}$:
\begin{equation}\label{Eq2:10}
       \small
       y_f = \max(\mathbf{y^a}) \cdot y_{ov}.
\end{equation}
Given the predicted action instance $\phi=\{\varphi_{c},\varphi_{w},C_a,y_f\}$ with refined boundary ($\varphi_{c},\varphi_{w}$), predicted action label $C_a$, and ranking score $y_f$, we employ the soft non-maximum suppression (soft-NMS)~\cite{Bodla:ICCV17} for post-processing.
In each iteration of soft-NMS, we represent the action instance with the maximum ranking score $y_{f_m}$ as $\phi_m$.
The ranking score $y_{f_k}$ of other instance $\phi_k$ will be decreased or not, according to the IoU computed with $\phi_m$:
\begin{equation}\label{Eq2:11}
\small
y'_{f_k} =
\begin{cases}
~~~y_{f_k}~~~~~~~~~~~~~~~~~~~~~~~~~~~~~,~~ \text{\footnotesize if~~$iou(\phi_k,\phi_m)< \rho$} \\
~~~y_{f_k} \cdot e^{-\frac{iou(\phi_k,\phi_m)^2}{\xi}}~~,~~ \text{\footnotesize if~~$iou(\phi_k,\phi_m) \geq \rho$}
\end{cases},
\end{equation}
where $\xi$ is the decay parameter and $\rho$ is the NMS threshold.

\section{Experiments}\label{sec:EX}
We empirically verify the merit of our GTAN by conducting the experiments of temporal action localization on two popular video recognition benchmarks, i.e., ActivityNet v1.3~\cite{ActivityNet} and THUMOS14~\cite{Thumos}.

\subsection{Datasets}
The \textbf{ActivityNet v1.3} dataset contains 19,994 videos in 200 classes collected from YouTube. The dataset is divided into three disjoint subsets: training, validation and testing, by 2:1:1. All the videos in the dataset have temporal annotations. The labels of testing set are not publicly available and the performances of action localization on ActivityNet dataset are reported on validation set. The \textbf{THUMOS14} dataset has 1,010 videos for validation and 1,574 videos for testing from 20 classes. Among all the videos, there are 220 and 212 videos with temporal annotations in validation and testing set, respectively. Following \cite{Xiong:ICCV17}, we train the model on validation set and perform evaluation on testing set.

\subsection{Experimental Settings}

\textbf{Implementations.} We utilize Pseudo-3D~\cite{Qiu:ICCV17} network as our 3D backbone. The network input is a 16-frame clip and the sample rate of frames is set as $8$. The 2,048-way outputs from pool5 layer are extracted as clip-level features. Table \ref{table3:1} summarizes the structures of 1D anchor layers. Moreover, we choose three temporal scale ratios $\{r_d\}_{d=1}^{3}=[2^0, 2^{1/3}, 2^{2/3}]$ derived from~\cite{LinYi:ICCV17}. The IoU threshold $\varepsilon$ in Gaussian grouping is set as $0.7$ by cross validation. The balancing parameters $\beta$ and $\gamma$ are also determined on a validation set and set as $2.0$ and $75$. $\xi$ and $\rho$ are set as $0.8$ and $0.75$ in soft-NMS. The parameter $\alpha_1$ and $\alpha_2$ are all set as $1.0$ by cross validation. We implement GTAN on Caffe~\cite{Yang:caffe} platform. In all the experiments, our networks are trained by utilizing stochastic gradient descent (SGD) with $0.9$ momentum. The initial learning rate is set as $0.001$, and decreased by $10\%$ after every $2.5k$ iterations on THUMOS14 and $10k$ iterations on ActivityNet. The mini-batch size is $16$ and the weight decay parameter is $0.0001$.

\begin{table}[!tb]
       \setlength{\belowcaptionskip}{-1pt}
       \centering
       \caption{\small The details of 1D temporal convolutional (anchor) layers. RF represents the size of receptive fields.}
       \vspace{0.02in}
       \scalebox{0.83}[0.83]{
              \begin{tabular}{{c|c|c|c|c|c}}
            \hline
            \text{id} & \text{type}  & \text{kernel size} &\text{\#channels} & \text{\#stride} & ~~\text{RF}~~  \\ \hhline{*{6}{-}}
            \text{1}  & \text{conv\_a1} & \text{3} &\text{512}  & \text{2} &  11   \\
            \text{2}  & \text{conv\_a2} & \text{3} &\text{512}  & \text{2} &  19   \\
            \text{3}  & \text{conv\_a3} & \text{3} &\text{1024} & \text{2} &  35   \\
            \text{4}  & \text{conv\_a4} & \text{3} &\text{1024} & \text{2} &  67   \\
            \text{5}  & \text{conv\_a5} & \text{3} &\text{2048} & \text{2} &  131  \\
            \text{6}  & \text{conv\_a6} & \text{3} &\text{2048} & \text{2} &  259  \\
            \text{7}  & \text{conv\_a7} & \text{3} &\text{4096} & \text{2} &  515  \\
            \text{8}  & \text{conv\_a8} & \text{3} &\text{4096} & \text{2} &  1027 \\ \hhline{*{6}{-}}
              \end{tabular}
     }
       \label{table3:1}
\vspace{-0.26in}
\end{table}

\textbf{Evaluation Metrics.} We follow the official evaluation metrics in each dataset for action detection task. On ActivityNet v1.3, the mean average precision (mAP) values with IoU thresholds between $0.5$ and $0.95$ (inclusive) with a step size $0.05$ are exploited for comparison. On THUMOS14, the mAP with IoU threshold $0.5$ is measured. We evaluate performances on top-100 and top-200 returned proposals in ActivityNet v1.3 and THUMOS14, respectively.

\subsection{Evaluation on Temporal Action Proposal}
We first examine the performances on temporal action proposal task, which is to only assess the boundary quality of action proposals, regardless of action classes. We compare the following advanced approaches: (1) Structure Segment Network (SSN) \cite{Xiong:ICCV17} generates action proposals by temporal actionness grouping. (2) Single Shot Action Detection (SSAD) \cite{Lin:MM17} is the 1D variant version of Single Shot Detection \cite{Liu:ECCV16}, which generates action proposals by multiple temporal anchor layers. (3) Convolution-De-Convolution Network (CDC) \cite{Shou:CVPR17} builds a 3D Conv-Deconv network to precisely localize the boundary of action instances at frame level. (4) Boundary Sensitive Network (BSN) \cite{Lin:ECCV18} locates temporal boundaries with three actionness curves and reranks proposals with neural networks. (5) Single Stream Temporal action proposal (SST) \cite{Buch:CVPR17} builds a RNN-based action proposal network, which could be implemented in a single stream over long video sequences to produce action proposals. (6) Complementary Temporal Action Proposal (CTAP)~\cite{Gao:ECCV18} balances the advantages and disadvantages between sliding window and actionness grouping approaches for final action proposal.

\begin{figure}[!tb]
       \centering
       \subfigure[]{\includegraphics[width=0.23\textwidth]{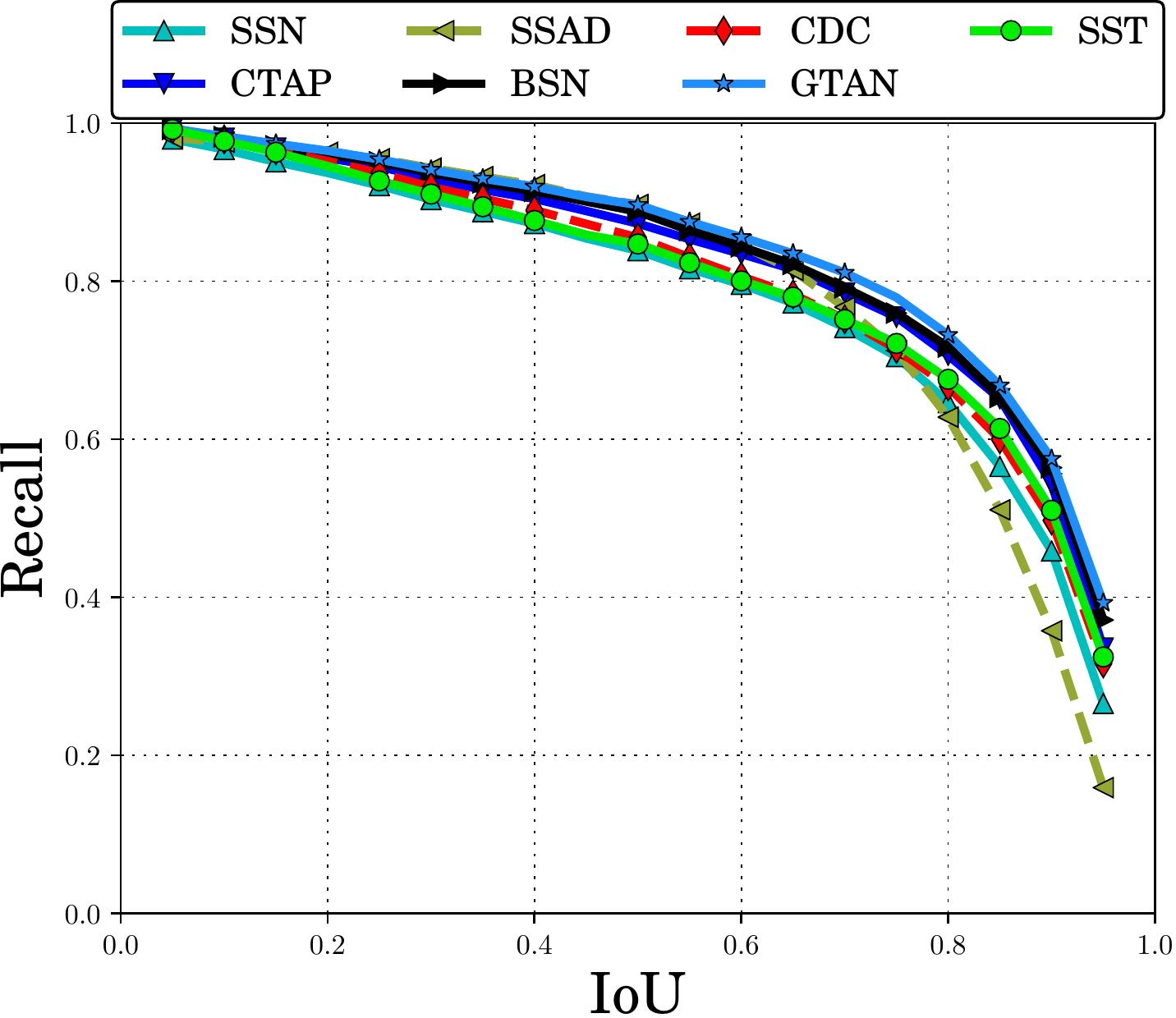}}
       \subfigure[]{\includegraphics[width=0.23\textwidth]{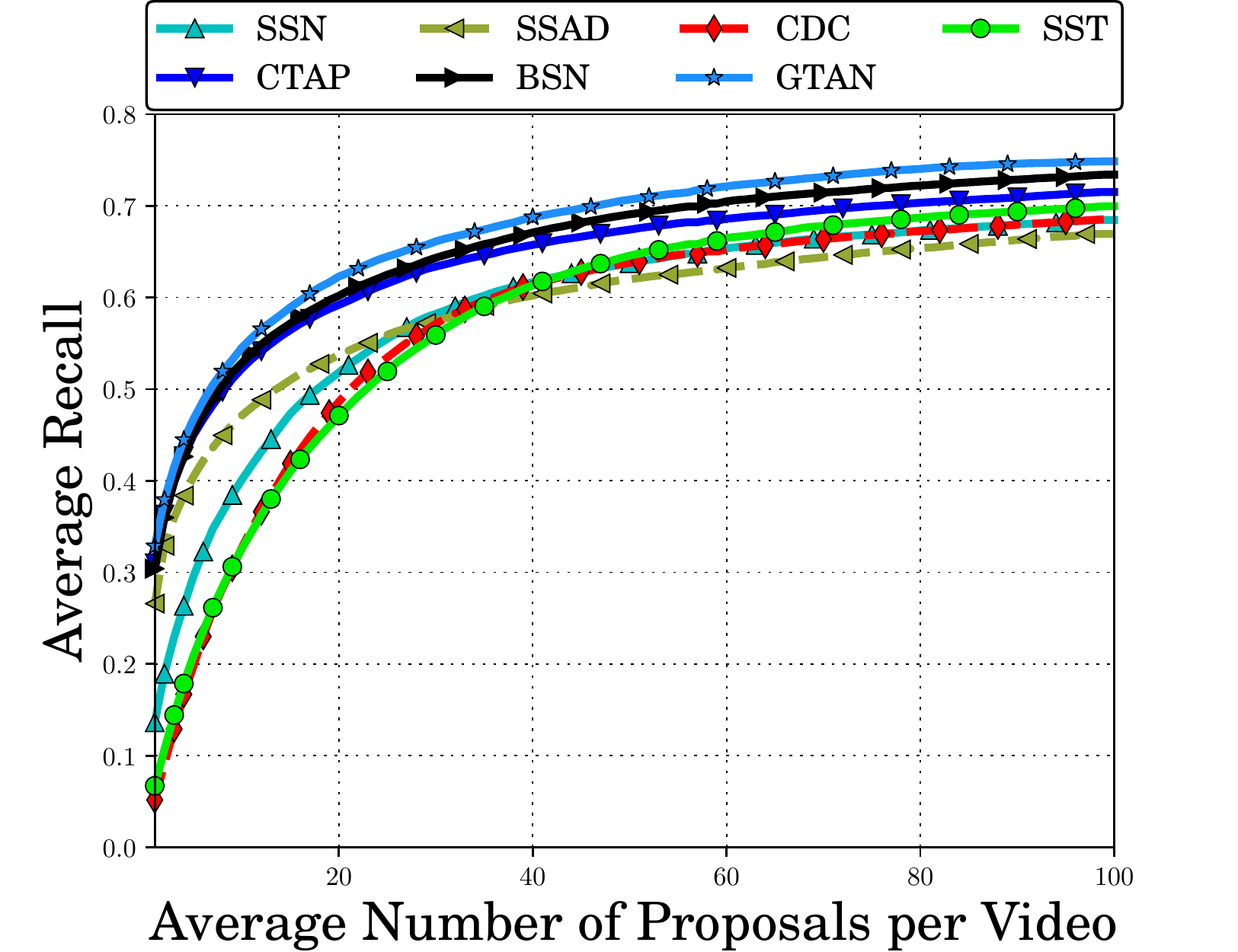}}
       \caption{\small (a) Recall-IoU and (b) AR-AN curve on ActivityNet.}
       \label{fig4:1}
       \vspace{-0.09in}
\end{figure}

\begin{table}[!tb]
       \setlength{\belowcaptionskip}{-1pt}
       \centering
       \caption{\small AR and AUC values on action proposal. IoU threshold: [0.5:0.05:1.0] for THUMOS14, [0.5:0.05:0.95] for ActivityNet.}
       \vspace{0.02in}
       \scalebox{0.80}[0.80]{
              \begin{tabular}{{|c|c|c@{~}c|c|}}
                     \hline
                     \multicolumn{1}{|c|}{\multirow{2}{*}{\text{Approach}}} & \multicolumn{1}{c|}{\text{THUMOS14}} & \multicolumn{2}{c|}{\text{ActivityNet}} & \multicolumn{1}{c|}{\text{ActivityNet (test server)}}\\ \hhline{*{1}{~}*{4}{-}}
                     \multicolumn{1}{|c|}{} & AR & ~~AR~~~ & AUC & AUC \\ \hhline{*{5}{-}}
                     \text{SST~\cite{Buch:CVPR17}}  & 37.9  & - & -& - \\
                     \text{~~CTAP~\cite{Gao:ECCV18}}& 50.1  & 73.2 & 65.7  & -   \\
                     \text{BSN~\cite{Lin:ECCV18}}   & 53.2  & 74.2 & 66.2  & 66.3 \\ \hhline{*{5}{-}}
                     \text{GTAN}                    & \textbf{54.3} & \textbf{74.8} & \textbf{67.1} & \textbf{67.4}  \\ \hhline{*{5}{-}}
              \end{tabular}
       }
    \label{table4:1}
    \vspace{-0.22in}
\end{table}

\begin{figure*}[!tb]
       \centering\includegraphics[width=0.91\textwidth]{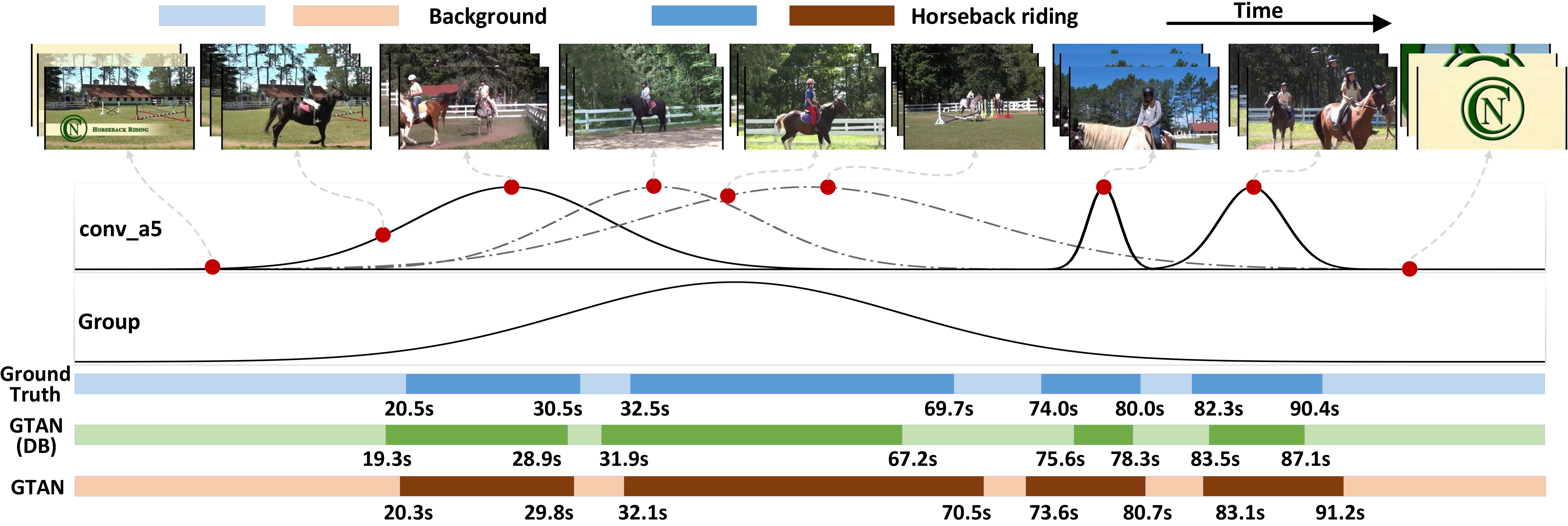}
       \caption{\small Visualization of action localization on a video example from ActivityNet by GTAN. The Gaussian kernels are learnt on the outputs of ``\text{conv\_a5}'' layer. The second and third kernels are mixed into a larger one. The default boxes (DB) are predicted by Gaussian~kernels.}
\label{fig4:3}
\vspace{-0.21in}
\end{figure*}

We adopt the standard metric of Average Recall in different IoU (AR) for action proposal on both datasets. Moreover, following the official evaluations in ActivityNet, we plot both Recall-IoU curve and Average Recall vs. Average Number of proposals per video (AR-AN) curve in Figure \ref{fig4:1}. In addition to AR metric, the area under AR-AN curve (AUC) is also reported in Table \ref{table4:1} as AUC is the measure on test server of ActivityNet. Overall, the performances across different metrics and two datasets consistently indicate that our GTAN leads to performance boost against baselines. In particular, AR of GTAN achieves 54.3\% and 74.8\% on THUMOS14 and ActivityNet respectively, making the absolute improvement over the best competitor BSN by 1.1\% and 0.6\%. GTAN surpasses BSN by 1.1\% in AUC when evaluating on online ActivityNet test server. The results demonstrate the advantages of exploiting temporal structure for localizing actions. Furthermore, as shown in Figure \ref{fig4:1}, the improvements are constantly attained across different IoU. In terms of AR-AN curve, GTAN also exhibits better performance on different number of top returned proposals. Even in the case when only less than 10 proposals are returned, GTAN still shows apparent improvements, indicating that GTAN is benefited from the mechanism of dynamically optimizing temporal scale of each proposal and the correct proposals are ranked at the~top.

\subsection{Evaluation on Gaussian Kernel and Grouping}

\begin{table}[!tb]
       \setlength{\belowcaptionskip}{-1pt}
       \centering
       \caption{\small Performance contribution of each design in GTAN.}
       \vspace{0.01in}
       \scalebox{0.76}[0.76]{
            \begin{tabular}{{|c|c|c| c|c | c|c|}}
            \hline
            \multicolumn{1}{|c|}{\multirow{1}{*}{\text{Approach}}} & \multicolumn{3}{c|}{\text{THUMOS14}} & \multicolumn{3}{c|}{\text{ActivityNet v1.3}} \\ \hhline{*{7}{-}}
            \text{Fixed Scale}         & \checkmark  &            &             & \checkmark  &            &              \\
            \text{Gaussian Kernel}     &             & \checkmark & \checkmark  &             & \checkmark & \checkmark   \\
            \text{Gaussian Grouping}   &             &            & \checkmark  &             &            & \checkmark   \\ \hhline{*{7}{-}}
            \text{mAP}                 &    33.5     &   37.1     &     38.2    &   29.8      &     31.6   &     34.3     \\ \hhline{*{7}{-}}
            \end{tabular}
       }
    \vspace{-0.0in}
    \label{table4:2}
    \vspace{-0.09in}
\end{table}

\begin{table}[!tb]
       \setlength{\belowcaptionskip}{-1pt}
       \centering
       \caption{\small The evaluations of Gaussian grouping on actions with different lengths. GTAN$^-$ excludes Gaussian grouping in GTAN.}
       \vspace{0.01in}
       \scalebox{0.84}[0.84]{
              \begin{tabular}{{|c|c|c|c|c|}}
                     \hline
                     \multicolumn{1}{|c|}{\multirow{2}{*}{\text{Approach}}} & \multicolumn{2}{c|}{\text{THUMOS14}} & \multicolumn{2}{c|}{\text{ActivityNet v1.3}} \\ \hhline{*{1}{~}*{4}{-}}
                     \multicolumn{1}{|c|}{}       & ~~$\geq 128$~~  & ~~~~\text{All}~~~~  & ~~$\geq 2048$~~ & ~~~~\text{All}~~~~       \\  \hhline{*{5}{-}}
                     \text{GTAN$^-$}              &       22.1      &     37.1       &     49.4   &       31.6            \\ \hhline{*{5}{-}}
                     \text{GTAN}                  &       25.9      &     38.2       &     54.2   &       34.3            \\ \hhline{*{5}{-}}
              \end{tabular}
       }
    \vspace{-0.0in}
    \label{table4:3}
    \vspace{-0.27in}
\end{table}

Next, we study how each design in GTAN influences the overall performance on temporal action localization task. Fixed Scale simply employs a fixed temporal interval for each cell or anchor in an anchor layer and such way is adopted in SSAD. Gaussian Kernel leverages the idea of learning one Gaussian kernel for each anchor to model temporal structure of an action and dynamically predict temporal scale of each action proposal. Gaussian Grouping further mixes multiple Gaussian kernels to characterize action proposals with various length. In the latter two cases, Gaussian pooling is utilized to augment the features of each anchor with contextual information.

Table \ref{table4:2} details the mAP performances by considering one more factor in GTAN on both datasets. Gaussian Kernel successfully boosts up the mAP performance from 33.5\% to 37.1\% and from 29.8\% to 31.6\% on THUMOS14 and ActivityNet v1.3, respectively. This somewhat reveals the weakness of Fixed Scale, where the temporal scale of each anchor is independent of temporal property of the action proposal. Gaussian Kernel, in comparison, models temporal structure and predicts a particular interval for each anchor on the fly. As such, the temporal localization or boundary of each action proposal is more accurate. Moreover, the features of each action proposal are simultaneously enhanced by contextual aggregation through Gaussian pooling and lead to better action classification. Gaussian grouping further contributes a mAP increase of 1.1\% and 2.7\%, respectively. The results verify the effectiveness and flexibility of mixing multiple Gaussian kernels to capture action proposals with arbitrary length. To better validate the impact of Gaussian grouping, we additionally evaluate GTAN on long action proposals. Here, we consider actions longer than 128 frames in THUMOS14 and 2048 frames in ActivityNet v1.3 as long actions, since the average duration of action instances in THUMOS14 is $\sim$4 seconds which is much smaller than that ($\sim$50 seconds) of ActivityNet. Table \ref{table4:3} shows the mAP comparisons between GTAN and its variant GTAN$^-$ which excludes Gaussian grouping. As expected, larger degree of improvement is attained on long action proposals by involving Gaussian grouping.

\begin{figure}[!tb]
       \centering
       \subfigure[]{\includegraphics[width=0.234\textwidth]{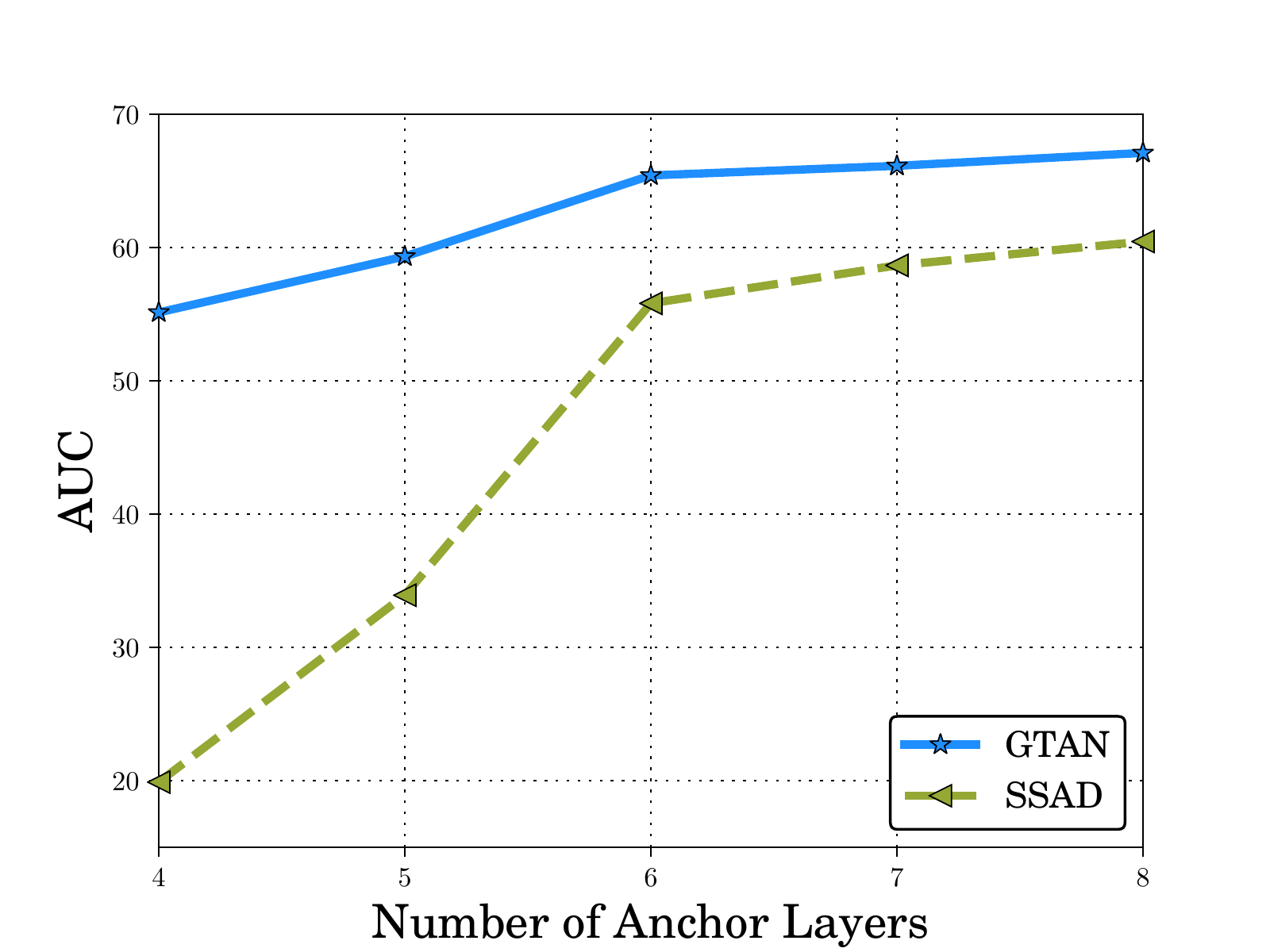}}
       \subfigure[]{\includegraphics[width=0.234\textwidth]{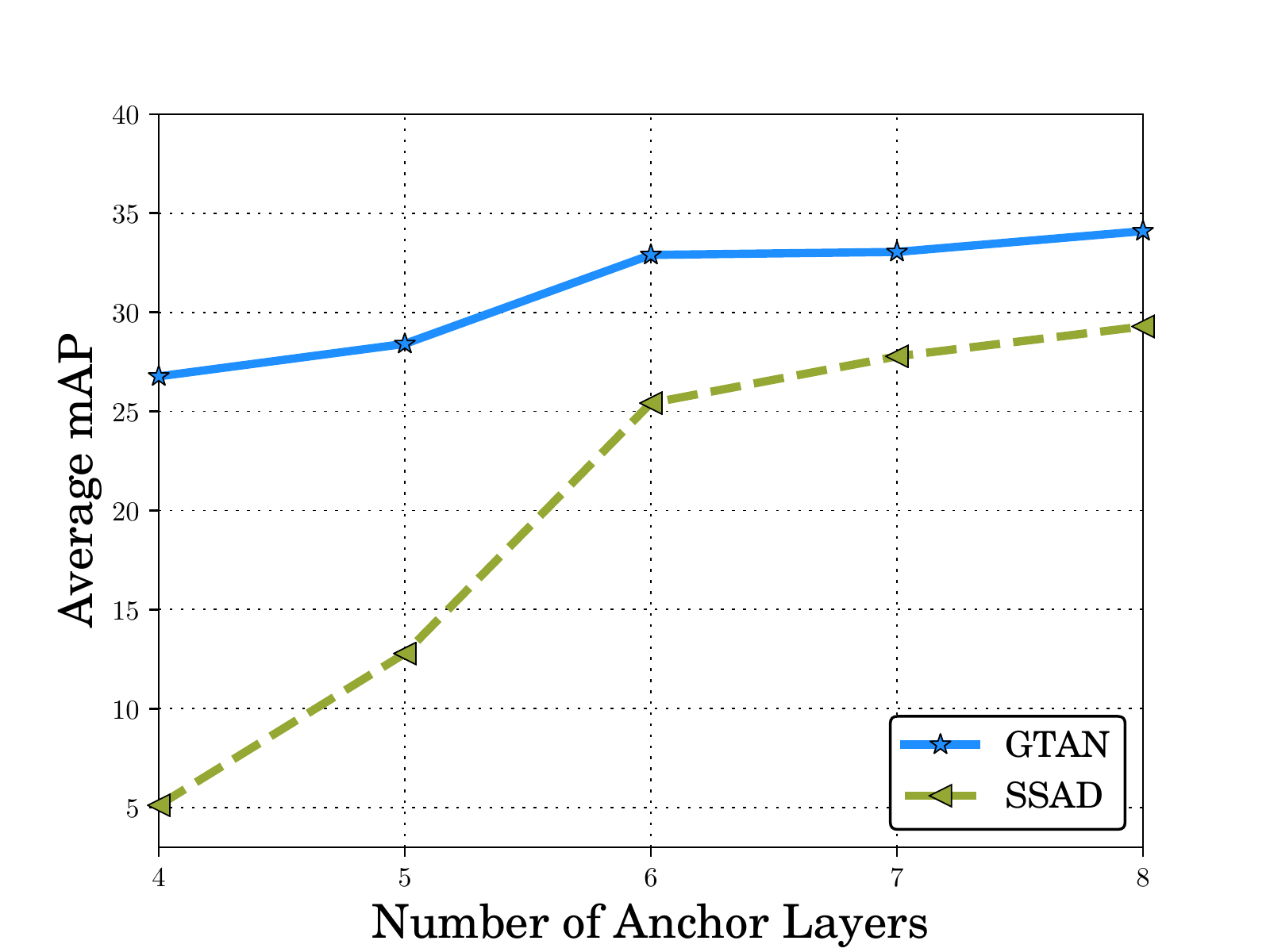}}
       \vspace{-0.05in}
       \caption{\small (a) AUC and (b) Average mAP performances of SSAD and GTAN with different number of anchor layers on temporal action proposal and localization tasks in ActivityNet.}
       \label{fig4:2}
    \vspace{-0.21in}
\end{figure}

\subsection{Evaluation on the Number of Anchor Layers}
In existing one-stage methods, e.g., SSAD, temporal scale is fixed in each anchor layer and the expansion of multiple temporal scales is implemented through increasing the number of anchor layers. Instead, our GTAN learns one Gaussian kernel for each anchor in every anchor layer and dynamically predicts temporal scale of the action proposal corresponding to each anchor. The grouping of multiple Gaussian kernels makes the temporal scale more flexible. Even with a small number of anchor layers, our GTAN should be more responsible to localize action proposals with various length in theory. Figure \ref{fig4:2} empirically compares the performances between SSAD and our GTAN on ActivityNet v1.3 when capitalizing on different number of anchor layers. As indicated by the results, GTAN consistently outperforms SSAD across different depths of anchor layers from 4 to 8 on both temporal action proposal and localization tasks. In general, more anchor layers provide better AUC and mAP performances. It is expected that the performance of SSAD decreases more sharply than that of GTAN when reducing the number of anchor layers. In the extreme case of 4 layers, GTAN still achieves 26.77\% in average mAP while SSAD only reaches 5.12\%, which again confirms the advantage of exploring temporal structure and predicting temporal scale of action proposals.

\subsection{Comparisons with State-of-the-Art}

\begin{table}[!tb]
       \setlength{\belowcaptionskip}{-1pt}
       \centering
       \caption{\small Performance comparisons of temporal action detection on THUMOS14, measured by mAP at different IoU thresholds $\alpha$.}
       \vspace{0.02in}
       \scalebox{0.76}[0.76]{
        \begin{tabular}{{c|c@{~~~~~~~~}c@{~~~~~~~~}c@{~~~~~~~~}c@{~~~~~~~~}c}}
        \hline
        \multicolumn{6}{c}{\multirow{1}{*}{\textbf{THUMOS14, mAP}@$\alpha$}} \\ \hhline{*{6}{-}}
        \multicolumn{1}{c|}{Approach} & 0.1 & 0.2 & 0.3 & 0.4 & 0.5\\ \hline \hline
        \multicolumn{6}{c}{\multirow{1}{*}{\text{Two-stage Action Localization}}}                         \\ \hline
        \text{Wang~\emph{et.al.}~\cite{Wang:thumos14}}              &  18.2 & 17.0 & 14.0 & 11.7 & 8.3    \\
        \text{FTP~\cite{Heilbron:CVPR16}}                           &  -    & -    & -    & -    & 13.5   \\
        \text{DAP~\cite{Escorcia:ECCV16}}                           &  -    & -    & -    & -    & 13.9   \\
        \text{Oneata~\emph{et.al.}~\cite{Oneata:thumos14}}          &  36.6 & 33.6 & 27.0 & 20.8 & 14.4   \\
        \text{Yuan~\emph{et.al.}~\cite{Yuan:CVPR16}}                &  51.4 & 42.6 & 33.6 & 26.1 & 18.8   \\
        \text{S-CNN~\cite{Shou:CVPR16}}                             &  47.7 & 43.5 & 36.3 & 28.7 & 19.0   \\
        \text{SST~\cite{Buch:CVPR17}}                               &  -    &  -   & 37.8 & -    & 23.0   \\
        \text{CDC~\cite{Shou:CVPR17}}                               &  -    &  -   & 40.1 & 29.4 & 23.3   \\
        \text{TURN~\cite{Gao:ICCV17}}                               &  54.0 & 50.9 & 44.1 & 34.9 & 25.6   \\
        \text{R-C3D~\cite{Xu:ICCV17}}                               &  54.5 & 51.5 & 44.8 & 35.6 & 28.9   \\
        \text{SSN~\cite{Xiong:ICCV17}}                              &  66.0 & 59.4 & 51.9 & 41.0 & 29.8   \\
        \text{CTAP~\cite{Gao:ECCV18}}                               &  -    &  -   &  -   &  -   & 29.9   \\
        \text{BSN~\cite{Lin:ECCV18}}                                &  -    &  -   & 53.5 & 45.0 & 36.9   \\ \hline
        \multicolumn{6}{c}{\multirow{1}{*}{\text{One-stage Action Localization}}}                         \\ \hline
        \text{Richard~\emph{et.al.}~\cite{Richard:CVPR16}}          &  39.7 & 35.7 & 30.0 & 23.2 & 15.2   \\
        \text{Yeung~\emph{et.al.}~\cite{Yeung:CVPR16}}              &  48.9 & 44.0 & 36.0 & 26.4 & 17.1   \\
        \text{SMS~\cite{Yuan:CVPR17}}                               &  51.0 & 45.2 & 36.5 & 27.8 & 17.8   \\
        \text{SSAD~\cite{Lin:MM17}}                                 &  50.1 & 47.8 & 43.0 & 35.0 & 24.6   \\
        \text{SS-TAD~\cite{Buch:BMVC17}}                            &  -    &  -   & 45.7 & -    & 29.2   \\ \hline\hline
        \text{GTAN (C3D)}                                           & 67.2  & 61.1 & 56.9 & 46.5 & 37.9   \\
        \text{GTAN}         &  \textbf{69.1}  &\textbf{63.7} & \textbf{57.8}  &  \textbf{47.2} &  \textbf{38.8} \\ \hhline{*{6}{-}}
         \end{tabular}
       }
    \label{table4:4}
    \vspace{-0.26in}
\end{table}

We compare with several state-of-the-art techniques on THUMOS14 and ActivityNet v1.3 datasets. Table \ref{table4:4} lists the mAP performances with different IoU thresholds on THUMOS14. For fair comparison, we additionally implement GTAN using C3D~\cite{Tran:ICCV15} as 3D ConvNet backbone. The results across different IoU values consistently indicate that GTAN exhibits better performance than others. In particular, the mAP@0.5 of GTAN achieve 37.9\% with C3D backbone, making the improvements over one-stage approaches SSAD and SS-TAD by 13.3\% and 8.7\%, which also employ C3D. Compared to the most advanced two-stage method BSN, our GTAN leads to 1.0\% and 1.9\% performance gains with C3D and P3D backbone, respectively. The superior results of GTAN demonstrate the advantages of modeling temporal structure of actions through Gaussian~kernel.

On ActivityNet v1.3, we summarize the performance comparisons on both validation and testing set in Table \ref{table4:5}. For testing set, we submitted the results of GTAN to online ActivityNet test server and evaluated the performance on the localization task. Similarly, GTAN surpasses the best competitor BSN by 0.6\% and 1.1\% on validation and testing set, respectively. Moreover, our one-stage GTAN is potentially simpler and faster than two-stage solutions, and tends to be more applicable to action localization in videos.

Figure \ref{fig4:3} showcases temporal localization results of one video from ActivityNet. The Gaussian kernels and grouping learnt on the outputs of ``\text{conv\_a5}'' layer are also visualized. As shown in the Figure, Gaussian kernels nicely capture the temporal structure of each action proposal and predict accurate default boxes for the final regression and classification.

\begin{table}[!tb]
       \setlength{\belowcaptionskip}{-1pt}
       \centering
       \caption{\small Comparisons of temporal action detection on ActivityNet.}
       \vspace{0.02in}
       \scalebox{0.76}[0.76]{
        \begin{tabular}{{c|c@{~~~~~}c@{~~~~}c@{~~~~}c@{~~}|c}}
        \hline
        \multicolumn{6}{c}{\multirow{1}{*}{\textbf{ActivityNet v1.3, mAP}}} \\ \hhline{*{6}{-}}
        \multicolumn{1}{c|}{\multirow{2}{*}{\text{Approach}}} & \multicolumn{4}{|c|}{\multirow{1}{*}{\text{validation}}} & \multicolumn{1}{|c}{\multirow{1}{*}{\text{testing}}} \\ \hhline{*{1}{~}*{5}{-}}
        \multicolumn{1}{c|}{}  & 0.5 & 0.75 & 0.95 & Average & Average\\ \hline \hline
        \text{Wang~\emph{et.al.}~\cite{Wang:anet16}}              &  45.11  & 4.11  & 0.05  & 16.41  & 14.62   \\
        \text{Singh~\emph{et.al.}~\cite{Singh:CVPR16}}            &  26.01  & 15.22 & 2.61  & 14.62  & 17.68   \\
        \text{Singh~\emph{et.al.}~\cite{Singh:ARXIV16}}           &  22.71  & 10.82 & 0.33  & 11.31  & 17.83   \\
        \text{CDC~\cite{Shou:CVPR17}}                             &  45.30  & 26.00 & 0.20  & 23.80  & 22.90   \\
        \text{TAG-D~\cite{Xiong:ARXIV17}}                         &  39.12  & 23.48 & 5.49  & 23.98  & 26.05   \\
        \text{SSN~\cite{Xiong:ICCV17}}                            &  -      &   -   &   -   &   -    & 28.28   \\
        \text{Lin~\emph{et.al.}~\cite{Lin:ARXIV17}}               &  48.99  & 32.91 & 7.87  & 32.26  & 33.40   \\
        \text{BSN~\cite{Lin:ECCV18}}                              &  52.50  & 33.53 & 8.85  & 33.72  & 34.42   \\ \hline\hline
        \text{GTAN}                                               &  \textbf{52.61} & \textbf{34.14} & \textbf{8.91}  &  \textbf{34.31} &  \textbf{35.54} \\ \hhline{*{6}{-}}
        \end{tabular}
       }
    \label{table4:5}
    \vspace{-0.26in}
\end{table}

\section{Conclusions}
We have presented Gaussian Temporal Awareness Networks (GTAN) which aim to explore temporal structure of actions for temporal action localization. Particularly, we study the problem of modeling temporal structure through learning a set of Gaussian kernels to dynamically predict temporal scale of each action proposal. To verify our claim, we have devised an one-stage action localization framework which measures one Gaussian kernel for each cell in every anchor layer. Multiple Gaussian kernels could be even mixed for the purpose of representing action proposals with various length. Another advantage of using Gaussian kernel is to enhance features of action proposals by leveraging contextual information through Gaussian pooling, which benefits the final regression and classification. Experiments conducted on two video datasets, i.e., THUMOS14 and ActivityNet v1.3, validate our proposal and analysis. Performance improvements are also observed when comparing to both one-stage and two-stage advanced techniques.

\textbf{Acknowledgments} This work was supported in part by the National Key R\&D Program of China under contract No. 2017YFB1002203 and NSFC No. 61872329.

{\small
\bibliographystyle{ieee_fullname}
\bibliography{egbib}
}

\end{document}